\documentclass[10pt,twocolumn,letterpaper]{article}

\usepackage{ijcb}
\usepackage{times}
\usepackage{epsfig}
\usepackage{graphicx}
\usepackage{amsmath}
\usepackage{amssymb}
\usepackage{pdflscape}
\usepackage{multirow}
\usepackage{graphicx}
\usepackage[accsupp]{axessibility}
\ijcbfinalcopy % *** Uncomment this line for the final submission

\ifijcbfinal\fi
\begin{document}

\title{Contactless Fingerprint Biometric Anti-Spoofing: An Unsupervised Deep Learning Approach}

%%%%%%%%% TITLE
% \title{Contactless Fingerprint Biometric Anti-Spoofing: An Unsupervised Deep Learning Approach with Attention Mechanism}

\author{Banafsheh Adami and Nima Karimian \\
West Virginia University\\
Computer Science and Electrical Engineering\\
{\tt\small ba00011@mix.wvu.edu, nima.karimian@mail.wvu.edu}}

\maketitle
\thispagestyle{empty}

%%%%%%%%% ABSTRACT
\begin{abstract}
Contactless fingerprint recognition offers a higher level of user comfort and addresses hygiene concerns more effectively. However, it is also more vulnerable to presentation attacks such as photo paper, paper-printout, and various display attacks, which makes it more challenging to implement in biometric systems compared to contact-based modalities. Limited research has been conducted on presentation attacks in contactless fingerprint systems, and these studies have encountered challenges in terms of generalization and scalability since both bonafide samples and presentation attacks are utilized during training model. Although this approach appears promising, it lacks the ability to handle unseen attacks, which is a crucial factor for developing PAD methods that can generalize effectively. We introduced an innovative anti-spoofing approach that combines an unsupervised autoencoder with a convolutional block attention module to address the limitations of existing methods. Our model is exclusively trained on bonafide images without exposure to any spoofed samples during the training phase. It is then evaluated against various types of presentation attack images in the testing phase. The scheme we proposed has achieved an average BPCER of 0.96\% with an APCER of 1.6\% for presentation attacks involving various types of spoofed samples.
\end{abstract}

%%%%%%%%% BODY TEXT
\section{Introduction}
\label{sec:intro}
Biometric systems have found extensive utility across various domains, including but not limited to law enforcement and forensics, singular identification, healthcare, and facilitating access control for smartphones and tablets. These applications contribute to enhanced convenience in our day-to-day activities. The demand for contactless biometric solutions is increasing rapidly due to hygiene-related issues. Fingerprints and facial biometrics are recognized as the primary modalities in the field of biometrics which extensive implementation by law enforcement agencies and national ID programs on a global scale~\cite{jain201650}. According to the biometric system market is projected to reach a value of \$82.9 billion by 2027~\cite{bworld}. Despite popularity of face authentication, it has encountered challenges during the pandemic, particularly regarding the use of face coverings~\cite{calbi2021consequences}, which prevents its high rate of 
detection~\cite{sun2013deep,yang2022mask,botezatu2022fun}.
Contactless fingerprint recognition provides great potential in various applications, offering a touchless and hygienic biometric authentication solution. Contactless fingerprinting is a cutting-edge technological advancement in the field of biometrics that eliminates the need for traditional bioscanner sensors~\cite{grosz2021c2cl,lin2018cnn,lin2018matching,kumar2013towards,cui2023monocular}. Instead, it relies solely on a smartphone camera lens for capturing and recording fingertip information. Compared to touch based fingerprint, it is considered to be more seamless and convenient and has a higher user acceptance.

\begin{figure*}
    \centering
    \includegraphics[width=0.95\linewidth]{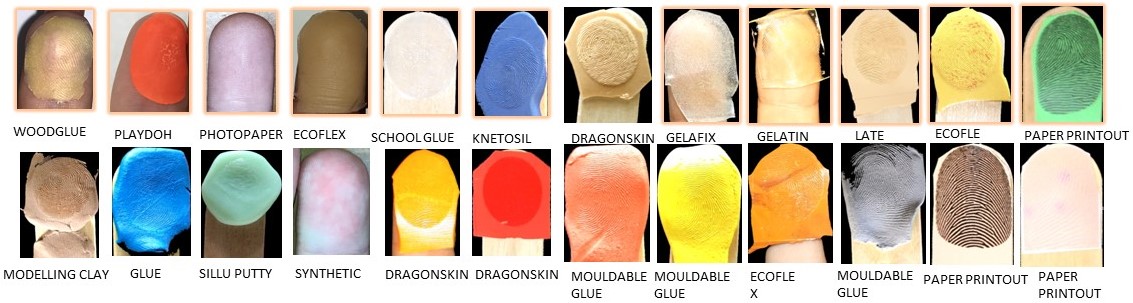}    
    \caption{Different spoofed samples from CLARKSON and COLFISPOOD datasets were employed in this paper. First four spoof samples are related to CLARKSON dataset (WOODGLUE,PLAYDOH, PHOTOPAPER, ECOFLEX), and others are spoof samples from COLFISPOOF dataset.}
    \label{fig:PAI}
\end{figure*}
Although contactless fingerprint technology provides convenient and widely accepted user experiences, it does come with various drawbacks. These include lower biometric performance, susceptibility to environmental influences, and vulnerabilities to presentation attacks~\cite{kolberg2023colfispoof}. Presentation attacks can compromise the security and reliability of biometric authentication systems, potentially leading to unauthorized access or identity theft~\cite{tolosana2019biometric}. Hence, developing an effective countermeasure against contactless fingerprint is crucial to detect and prevent any unseen presentation attacks. Contactless fingerprint systems used smartphones for capturing photo-based finger images are more vulnerable to spoofing due to the use of a single type of camera and limited computational capabilities (Figure.~\ref{fig:PAI} shows fingerprint spoof images fabricated for this study). While well-studied has been conducted on contact-based fingerprint recognition and its vulnerabilities to presentation attacks, limited attention has been given to studying contactless fingerprint presentation attacks. In recent years, only few numbers of approaches have been explored to identify various contactless fingerprint presentation attacks using hand-crafted features, and deeply learned features. Despite significant progress in contactless fingerprint PAD, there are still limitations associated with existing methods. These limitations include:

\begin{itemize}
  \item Disparities in data distributions: existing PAD approaches on contactless fingerprint assume similar data distributions between training and testing scenarios. However, this assumption leads to limited generalization capabilities of PAD methods when faced with real-world situations, especially with unseen attacks.
  \item Multiple types of presentation attacks: contactless fingerprint presentation attacks can take various forms, including printed attack, latex, ecoflex, and more. Thus, generating and creating a labeled training set that covers all possible presentation attacks for each new application scenario is impractical.
\end{itemize}

\begin{table*}[htp]
\begin{center}
\resizebox{\textwidth}{!}{%
\begin{tabular}{|l|l|l|l|l|l|l|}

\hline
Author & Year & Method & Database & Spoof type  & Results \\ \hline
Tanej et al.~\cite{taneja2016fingerphoto} & 2016 & Hand crafted & \begin{tabular}[c]{@{}l@{}}IIITD: \\ class: 128 \\ images: 5100 \end{tabular} & \begin{tabular}[c]{@{}l@{}}Print Attack\\ Photo Attack\end{tabular}   & EER = 3.71\% \\ \hline
Wasnik et al.~\cite{wasnik2018presentation} & 2018 &\begin{tabular}[c]{@{}l@{}} Hand crafted LBP,\\ BSIF, HOG, SVM \end{tabular}& \begin{tabular}[c]{@{}l@{}} subjects: 50 \\images: 250 \\  videos: 150\end{tabular}  & \begin{tabular}[c]{@{}l@{}}print artefact\\ electronic replay\\ elctronic display\end{tabular}  & \begin{tabular}[c]{@{}l@{}}BPCER = 1.8, 0, 0.66,\\ APCER = 10\end{tabular} \\ \hline
Fujito et al.~\cite{fujio2018face}& 2018 & AlexNet & \begin{tabular}[c]{@{}l@{}}Live: 4096\\ spoofe sample: 8192\end{tabular} &  \begin{tabular}[c]{@{}l@{}}Print Attack \\ Photo Attack\end{tabular}  & HTER = 0.04\% \\ \hline
Marasco et al.~\cite{marasco2021fingerphoto}, \cite{marasco2021deep} & 2022 & \begin{tabular}[c]{@{}l@{}}AlexNet DenseNet201,\\  ResNet18,DenseNet121,\\ ResNet34, MobileNEt-V2\end{tabular} & IIITD  &\begin{tabular}[c]{@{}l@{}} Print Attack\\Photo Attack\end{tabular} &  \begin{tabular}[c]{@{}l@{}}D-EER\_AlexNet = 2.14\\ D-EER\_ResNet = 0.96\%\end{tabular} \\ \hline
Kolberg et al.~\cite{kolberg2023colfispoof}& 2023 & Not Reported & \begin{tabular}[c]{@{}l@{}}COLFISPOOF:\\ 7200 spoof samples \\ 72 different PAI\end{tabular} & Not Reported & \begin{tabular}[c]{@{}l@{}}Knetosil, Mould glue, \\ latex, silly putty,\\ paper printout, s\\ chool glue, \\ dragonskin, \\ ecoflex, gelatin, \\ glue,  modelling clau, \\ playdoh\end{tabular}   \\ \hline
Purnapatra et al.~\cite{purnapatra2023presentation} & 2023 & \begin{tabular}[c]{@{}l@{}}DenseNet 121,\\ NASNet\end{tabular} & \begin{tabular}[c]{@{}l@{}}35 subjects with 12 devices\\ attack sample: 7548 \\ synthetic: 10000\end{tabular}  & \begin{tabular}[c]{@{}l@{}}ecoflex, playdoh,\\ wood glue, \\ synthetic, fingerphoto,\\ latex\end{tabular}  &\begin{tabular}[c]{@{}l@{}} APCER = 0.14\% \\ BPCER = 0.18\% \end{tabular}\\ 
\hline

Hailin Li et al.\cite{li2023deep} & 2023 & \begin{tabular}[c]{@{}l@{}}AlexNet,DenseNet201,\\MobileNet-V2,ResNet50 \\ NasNet, GoogleNet,\\EfficientNEt-B0\\Vision Transformer\end{tabular} & \begin{tabular}[c]{@{}l@{}}5886 bonafide\\ 4247 attack sample\\ four PAIs types\end{tabular}  & \begin{tabular}[c]{@{}l@{}}ecoflex, playdoh,\\ wood glue, \\ synthetic, fingerphoto, \\latex\end{tabular}  &\begin{tabular}[c]{@{}l@{}} They report APCER and BPCER\\in 4 cases,\\in each case one PAI\\used only for testing \\and three remains \\used for training \end{tabular}\\ \hline

Puranpatra et.al.\cite{purnapatra2023liveness} & 2023 & \begin{tabular}[c]{@{}l@{}}Combination of \\two CNN\end{tabular} & \begin{tabular}[c]{@{}l@{}}5886 bonafide\\ 4247 attack sample\\ four PAI types \end{tabular}  & \begin{tabular}[c]{@{}l@{}}ecoflex, playdoh,\\ wood glue, \\  fingerphoto, \\latex\end{tabular}  &\begin{tabular}[c]{@{}l@{}} BPCER = 0.62\\ APCER = 11.35\\ ACER = 6 \end{tabular}\\ \hline

B Adami et al.\cite{adami2023universal} & 2023 & \begin{tabular}[c]{@{}l@{}}Resnet-18/LeakyRelu,\\(Combined Loss)\end{tabular} & \begin{tabular}[c]{@{}l@{}}5886 bonafide\\ 4247 attack sample\\ 10,000 synthetic \end{tabular}  & \begin{tabular}[c]{@{}l@{}}ecoflex, playdoh,\\ wood glue, \\ synthetic, fingerphoto, \\latex\end{tabular}  &\begin{tabular}[c]{@{}l@{}} BPCER = 0.12\\ APCER = 0.63\\ ACER = 0.68 \end{tabular}\\ \hline

Our work & 2024 & \begin{tabular}[c]{@{}l@{}}convolution autoencoder,\\ \textbf{CBAM-autoencoder}\\
\textbf{(proposed work)},\\Swin-transformer\end{tabular} & \begin{tabular}[c]{@{}l@{}}35 subjects with 12 devices\\ attack sample: 7548 \\ synthetic: 10000\end{tabular}  & \begin{tabular}[c]{@{}l@{}}ecoflex, playdoh, wood glue, \\ synthetic, fingerphoto, latex\end{tabular}  &\begin{tabular}[c]{@{}l@{}} APCER = 1.6\% \\ BPCER = 0.96\% \end{tabular}\\ \hline

\end{tabular}%
}

\caption{Summary of previous works for contactless fingerprint anti-spoofing. HOG-- histogram of oriented gradients (HOG), SVM-- support vector machine, LBP--local binary patterns, BSIF--binarized statistical image features, EER -- equal error rate, TAR -- true acceptance rate, FAR -- false acceptance rate BPCER--bonafide presentation classification error rate, HTER -- half total error rate, APCER-- attack presentation classification error rate.}
\label{tab:previous works}
\end{center}
\end{table*}

In this paper, we have conducted generalized PAD and introduced a novel end-to-end trainable PAD approach called Unsupervised deep auto-encoder to detect the unlabeled and unseen presentation attack and build a robust PAD model. The main contributions of this paper are summarized:
\begin{itemize}
    \item A novel unsupervised convolutional block attention module auto-encoder approach that is able to leverage bonafide (live data) data to build robust PAD model which is independent from unseen spoofed data. 
    \item  Developing advanced deep learning architecture based on live image and test against unseen spoofed samples from two public database to improve on the existing methods in noncontact-fingerprint PAD.
    \item Developing an deep architecture which is able to successfully classify unseen live subject as live samples. 
    \item  Evaluating the performance of Presentation Attack Detection (PAD) in several contactless fingerprint spoofed database tests and compared it to a supervised learning approach where the detection of unseen attacks was not possible. 
    
\end{itemize}

The paper organizes as follows. Section~\ref{sec:lite} discusses the previous works and Section~\ref{sec:Proposed} discusses potential attacks while introducing our system environment. Section~\ref{sec:Proposed} presents our new unsupervised deep learning architecture that can detect any unseen attack. Section~\ref{sec:Setup} shows the experimental setup and results. The comparison of proposed work with existing literature has been demonstrated in Section \ref{sec:Setup}. Finally, we conclude this paper in Section~\ref{sec:conclude}. 

\section{Related work}
\label{sec:lite} 
As mentioned earlier, only a few studies have been conducted to develop a contactless fingerprint PAD (Presentation Attack Detection) system. Table.~\ref{tab:previous works} provides a concise summary of the previous works related to contactless fingerprint technology. Fujio et al.~\cite{fujio2018face} were among the pioneers in exploring the application of deep neural networks for contactless fingerprint anti-spoofing. They achieved an impressive half-error rate of only 0.04\%. Marasco et al.~\cite{marasco2021fingerphoto} employed Convolutional Neural Network (CNN) architectures like ResNet and AlexNet on the IIITD Spoofed Finger Photo Database. They achieved a Detection Equal Error Rate (D-EER) of 2.14\% for AlexNet and 0.96\% for ResNet. Subsequently, they made slight improvements in comparison to the baseline approach\cite{marasco2021deep}. Despite the promising D-EER results for the ResNet architecture, it's important to note that the model was trained on both live and spoofed images, which might not be fully representative of real-world scenarios and could impact scalability. Furthermore, in 2022, they introduced a method to enhance the PAD system's robustness against color paper print-out attacks~\cite{marasco2022late}. Their proposed framework involves segmenting input photos using a U-Net with a ResNet-50 backbone. Minutiae points extracted from the images are projected into multiple color spaces, and patches around these points are generated to enhance local texture information. Their efforts led to an impressive achievement of APCER (Attack Presentation Classification Error Rate) at 0.1\% when BPCER (bonafide Presentation Classification Error Rate) was 2.67\%. Kolberg et al.~\cite{kolberg2023colfispoof} introduced the COLFISPOOF dataset specifically designed for non-contact fingerprint Presentation Attack Detection (PAD) purposes. This comprehensive dataset comprises 7200 samples, covering 72 distinct types of spoofed attacks, all captured using two different smartphone devices. Purnapatra et al.~\cite{purnapatra2023presentation} introduced the utilization of DenseNet-121 and NasNetMobile models in conjunction with a newly accessible public database. They incorporated both live and spoof data in their training process and achieved an APCER of 0.14\% and a BPCER of 0.18\%.
In the study by Hailin Li et al.~\cite{li2023deep}, the effectiveness of presentation attack detection (PAD) was showcased using several models, including AlexNet, DenseNet-201, MobileNet-V2, NASNet, ResNet50, and Vision Transformer. The vision transformer achieves the best APCER and BPCER among other methods, which were previously used in many fields, such as image compression~\cite{khoshkhahtinat2023multi}. The research encompassed a comprehensive approach, involving over 5,886 genuine samples and 4,247 spoof samples. Four distinct training cases were considered, each focusing on a single type of spoof for testing (ecoflex, photo paper, playdoh, and wood glue). Notably, the ResNet50 model achieved an 8.6\% equal error rate (EER). Despite the promising performance exhibited by these recent models in recognizing fraudulent images during training, their ability to generalize to new counterfeit images proved limited, resulting in suboptimal performance in such scenarios.
Puranpatra et al.~\cite{purnapatra2023liveness} held a competition on fingerprint liveness detection. The winner of competiton achieve APCER=9.20\% for paper printout, APCER=0\% for ecoflex, playdoh and latex, APCER = 0.1\% for woodglue and APCER=99.9\% for synthetic fingertip at BPCER=0.62\%.

\section{Proposed Method}
\label{sec:Proposed}
% As illustrated in Figure.~\ref{fig:model}, we developed an attention based unsupervised method for contactless fingerprint anti-spoofing purpose. The convolutional autoencoder (CAE) contains symmetric encoder and decoder~\cite{goodfellow2016deep}. The convolutional Block Attention Module (CBAM) has Incorporated into a convolutional autoencoder to achieve a better spoof detection. The convolutional block attention module consists of two attention mechanisms: the channel attention module (CAM) and the spatial attention module (SAM). The channel attention module captures interdependencies between feature channels of input live finger images, while the spatial attention module focuses on capturing spatial relationships within feature maps of live input finger images~\cite{woo2018cbam}.

\begin{figure*}
    \centering
    \includegraphics[width=0.95\linewidth]{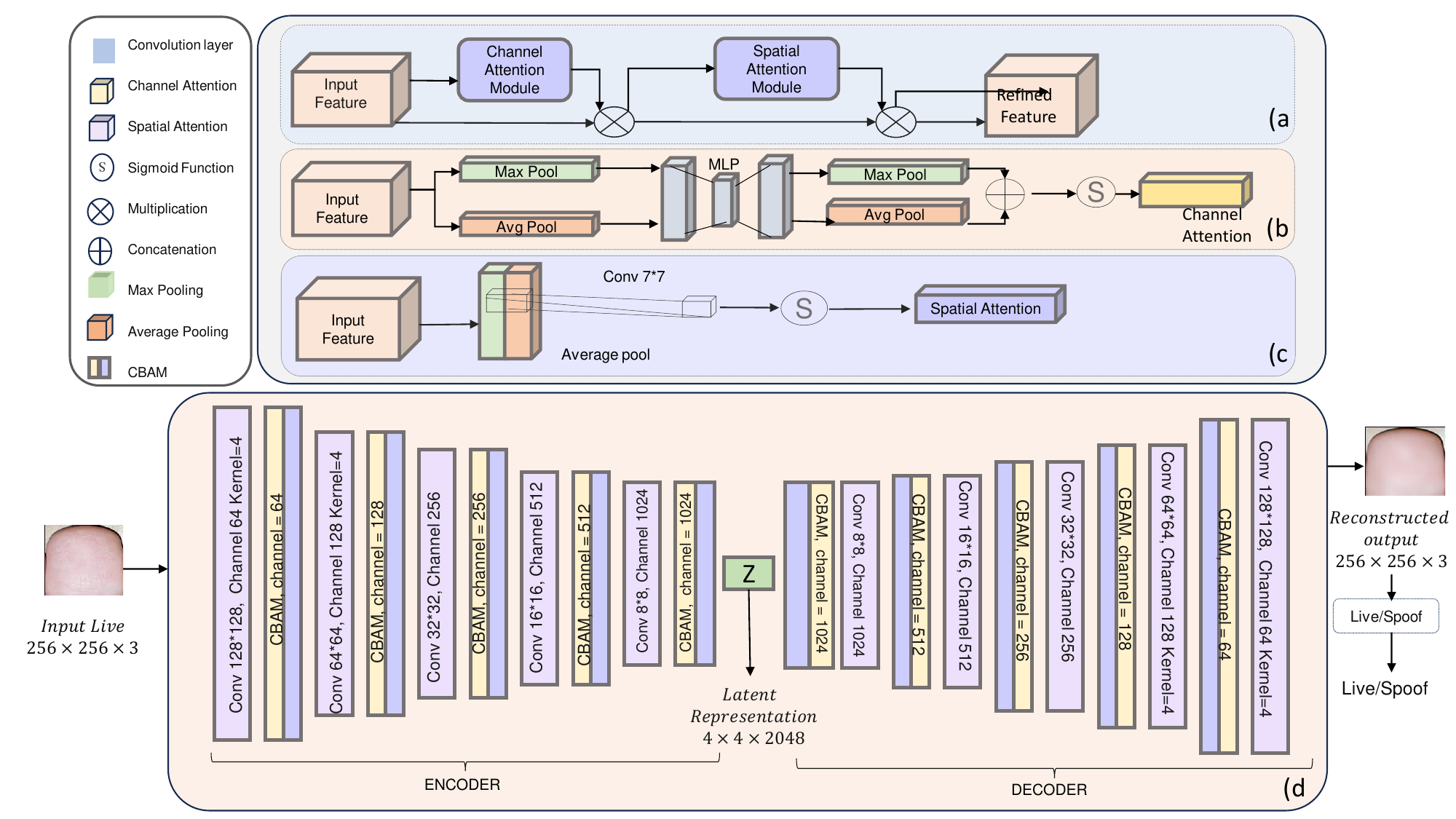}    
    \caption{proposed convolutional block attention module autoencoder. a) convolutional block attention module (CBAM). b) channel attention module (CAM), c) spatial attention module (SAM), d) convolutional autoencoder with attention mechanism.}
    \label{fig:model}
\end{figure*}

According to the Figure.~\ref{fig:model}, the proposed method aims to enhance feature representation of live samples by incorporating attention modules into the convolutional autoencoder (CAE). The CAE contains symmetric encoder and decoder~\cite{goodfellow2016deep}. The convolutional block attention module (CBAM) has incorporated into a CAE to achieve a better spoof detection.

Convolutional layers in autoencoder efficiently learn spatial representation and extract meaningful features from live finger images~\cite{bedi2021plant}~\cite{zhang2018better}~\cite{michelucci2022introduction}. We enhance the classification performance of unsupervised model by incorporating attention mechanism into convolutional autoencoder~\cite{guo2022attention} which has been used previously in many fields~\cite{obeso2022visual}~\cite{farahani2023time}. 
The CBAM consists of two attention mechanisms: the channel attention module (CAM) and the spatial attention module (SAM).

The channel attention module captures inter dependencies between feature channels of input live finger images and calculates as follows:
\begin{align}
        & M_{channel}(X) = \sigma(MLP(AvgPool(X)) \\
        &+ MLP(MaxPool(X))),
\end{align}
where $\sigma$ is sigmoid activation function, $MLP$ is multi-layer perceptron.  
The spatial attention module focuses on capturing the spatial relationships within the feature maps of live input finger images~\cite{woo2018cbam}. The spatial attention module, as shown in Figure~\ref{fig:model}-c, utilizes the spatial relationships of live features to determine the most informative parts of the live input features. This enhancement aims to improve the model's performance in classification tasks. The module enables the model to focus on various spatial regions within the feature maps, thereby concentrating on the informative regions of the input live features~\cite{woo2018cbam}. By attending to relevant spatial locations of live features, the model can emphasize fine-grained details and capture spatial relationships more effectively~\cite{chen2017sca}. To calculate the spatial attention of the live input feature, we perform average-pooling and max-pooling operations along the channel dimension of the live feature map. This generates a concise and informative representation of the features, which are then concatenated. After concatenating the average-pooling and max-pooling results, we apply convolution layers ($f^{7\times7}$) to build a spatial attention map ($M_{s}\in \mathbf{R}^{1\times H \times W}$). 
This spatial attention mechanism aids in preserving spatial information during the encoding and decoding processes. Equation \ref{eq:spatial} shows how we calculate spatial attention of live features: 
\begin{align}
    M_{spatial}(X)=\sigma (f^{7\times7} [AvgPool(X); MaxPool(X)]),
    \label{eq:spatial}
\end{align}

This integration enables the capture of both channel-wise and spatial-wise attention, facilitating the learning of representations for live finger images.

We applied CAM and SAM in a sequential manner~\cite{woo2018cbam}. By separating the channel attention map and spatial attention map of the live finger features, we enable a greater emphasis on the significant and salient features within the input live features. Furthermore, in the attention module, we have integrated global pooling to harness comprehensive global spatial information~\cite{woo2018cbam}~\cite{guo2022attention}.

Assume $X$ is an input live finger feature, and $X \in \mathbf{R}^{C\times H\times W }$, it deduce 1D channel attention map ($M_{c}\in \mathbf{R}^{C\times 1 \times 1}$), and 2D spatial channel attention map ($M_{s}\in \mathbf{R}^{1\times H \times W}$), where $C$ is channel, $H$ is height, and $W$ is width. Figure.~\ref{fig:model}-a, shows the whole attention process, and we can summarize them to below equations:
\begin{align}
    X^{\prime} = M_{c}(X)\otimes(X),
\end{align}
\begin{align}
    X^{\prime\prime} = M_{s}(X^{\prime})\otimes (X^{\prime}),
\end{align}

where, $M_{c}$ and $M_{s}$ are channel and spatial attention module. $X^\prime$ and $X^{\prime\prime}$ are refined output after applying channel and spatial attention module, and $\otimes$ is hadamard multiplication.

Figure.~\ref{fig:model} shows the proposed attention-based CAE model. As shows in Figure.~\ref{fig:model}, we apply channel attention module and spatial attention module after each convolution layer in both encoder and decoder to better capture and exploit the inherent structure and dependencies present in the input live features. 
The CBAM module with integrating attention mechanisms to autoencoder, enabled the network to simultaneously focus on spatially important regions and channel-wise features.
Ultimately, the proposed attention-based convolutional autoencoder model is an unsupervised approach and it trained only on live samples and becomes capable of distinguishing between live and spoof finger images.
The proposed method tries to minimize the mean squared error (MSE), which measures the pixel-wise difference between the reconstructed live image and the original input live image ~\cite{zagoruyko2016paying}. The reconstruction error can provide valuable insights into the quality of reconstructed output and latent space representation.

\subsection{Implementation Details}
\label{sec:detail}
The proposed method follows an unsupervised training approach, utilizing a dataset of unlabeled live images to train the attention based CAE model. The unsupervised nature of the training allows the model to learn meaningful representations without requiring explicit class labels. We only use live data as input images and feed them into our proposed model. According to Equation~\ref{eq:conv}, after each convolutional layer, we applied a rectified linear unit (Relu activation function) and then pass the output of the convolutional layer to the CAM (Equation~\ref{eq:convchannel}) followed by spatial attention (Equation~\ref{eq:convspatial}). As can be seen in Figure.~\ref{fig:model}, the proposed encoder and decoder consist of 5 convolutional layers and 5 spatial and channel attention layers with a kernel size of 4. The idea behind the small kernel size is to better capture temporal features from input data. After each CBAM we apply a dropout layer with the rate of 0.5. After each CBAM, we apply a dropout layer with a rate of 0.5. The idea behind applying the dropout layer in the proposed method is to prevent overfitting and generalize better to unseen types of spoof attacks. Equation ~\ref{eq:conv},~\ref{eq:convchannel},~\ref{eq:convspatial} shows how our algorithm works. We follow this steps in both encoder and decoder part of attention based convolutional autoencoder to find the best refined output and minimize the reconstruction loss of live input images during training. 

\begin{align}
    Conv(x ^{(C \times H \times W)})  \rightarrow  x ^{(C^\prime \times H^\prime \times W^\prime)},
    \label{eq:conv}
\end{align}

\begin{equation}
x'^{(C^\prime \times H^\prime \times W^\prime)} \rightarrow M_{c} ( x^{(C^\prime \times H^\prime \times W^\prime)}) \otimes x^{(C^\prime \times H^\prime \times W^\prime)},
\label{eq:convchannel}
\end{equation}

\begin{equation}
x''^{(C^\prime \times H^\prime \times W^\prime)} \rightarrow M_{s} ( x'^{(C^\prime \times H^\prime \times W^\prime)}) \otimes x'^{(C^\prime \times H^\prime \times W^\prime)},
\label{eq:convspatial}
\end{equation}

In equations \ref{eq:conv}, \ref{eq:convchannel} and \ref{eq:convspatial} values are as follows: $X$, $C$, $H$, $W$ are input feature, channel, height and width. $x^\prime$, $C^\prime$, $H^\prime$, $W^\prime$ are output feature, channel, height and width after applying channel attention module. $x''$, $C''$, $H''$, $W''$ are output feature, channel, height, and width after applying spatial attention module. Lastly, $\otimes$ is hadamard multiplication.

Once we obtain the reconstructed image, we calculate the MSE loss value (See Equation~\ref{eq:loss}).
\begin{align}
    Min L(x) = \frac{1}{N}\sum_{i=1}^{N}\left\| X_{i} - \hat{X}_{i} \right\|^{2},
    \label{eq:loss}
\end{align}
In above equation, $L$ is the MSE loss value, $N$ is the total number of samples, $X_{i}$ is the reconstruction value for live images and $\hat{X}_{i}$ is the reconstruction value for test images (live/spoof). 
The MSE provides a measure of how well the proposed model is able to reconstruct the input image. A lower MSE indicates better reconstruction, as it signifies a smaller average difference between the original live and reconstructed live pixel values. The reconstructing error is considered as the classification threshold to classify images as live or spoof. During unsupervised training, the model minimizes the mean squared error (MSE) loss, which measures the pixel-wise difference between the reconstructed image and the original input image. The Adam optimizer is employed to optimize the model's parameters.

\section{Experimental Setup}
\label{sec:Setup}

\subsection{Database}
In our study, we utilized three public database including CLARKSON~\cite{purnapatra2023presentation}, COLFISPOOF~\cite{kolberg2023colfispoof}, and IIITD Spoofed Fingerphoto Database \cite{sankaran2015smartphone,taneja2016fingerphoto}. The CLARKSON  dataset consists of 7,500 images of four-finger attacks, along with over 14,000 manually segmented images of single-fingertip attacks. Additionally, there are 10,000 synthetic fingertip images created using deepfake techniques. The dataset was gathered from six different Presentation Attack Instruments (PAI) spanning three levels of difficulty. Furthermore, the CLARKSON database comprises a total of 31,702 images of 26 subjects recorded from live finger photo. Among them, 2,150 images were collected from four-finger scenario, while  7,768 collected from single fingertip. The evaluation each device's effectiveness and performance involved the use of six different smartphone: iPhone x, iPhone 7, Samsung Galaxy S9, Google Pixel, Samsung Galaxy S6, and S7. For the spoofed image, different smartphone is utilized to generate spoofed fingertip such as synthetic, Ecoflex PAI, Playdoh PAI, Wood Glue PAI, Finger Photo PAI, and Latex PAI. In contrast, COLFISPOOF~\cite{kolberg2023colfispoof} database contains only spoof images from different categories including dragonskin, ecoflex, gelafix, gelatin, glue, knetosil, latex, modelling-clay, moduldable-glue, paper-printout, playdoh, and silly-putty. Table.~\ref{clarksondataset} and ~\ref{COLFISPOOFDATASET} shows the statistics of the databases. The IIITD contains images of spoofed fingerphotos. These images were captured using devices like the OnePlus One and Nokia phones, involved devices such as the iPad, Laptop, Nexus, and printouts. The dataset comprises two categories of fake samples: photo paper and printed paper spoofs, obtained by associating 64 subjects with two distinct fingers each where 2,048 images are print attacks, and 6,144 images are for photo attacks. Additionally, the images were taken under varying lighting conditions and against two different background variations. \textit{Note that, CLARKSON dataset in this study contains less number of samples both live and spoofed compared the one reported in~\ref{clarksondataset} for the purpose of competition. The original database has synthetic spoofed sample which has not been investigated in this study. Based on the results reported in~\ref{clarksondataset}, PAD for synthetic data is not difficult compared to photo paper.}

\begin{table}[htbp]
\centering
\begin{tabular}{cc}
\hline
\multicolumn{2}{c}{SPOOF} \\
\hline
API & NUMBER of IMAGES \\
\hline
ECOFLEX & 1248 \\
PHOTOPAPER & 1104 \\
PLAYDOH & 1700 \\
WOODGLUE & 272 \\
\hline
\multicolumn{2}{c}{LIVE (26 subjects)} \\
\hline
LIVE & 5886 \\
\hline
\end{tabular}
\caption{Statistics of CLARKSON dataset\cite{purnapatra2023presentation}}
\label{clarksondataset}
\end{table}

\begin{table}[htbp]
\centering
\begin{tabular}{c c}
\hline
\textbf{Spoof} & \textbf{Number of Images} \\
\hline
DRAGONSKIN & 1700 \\

ECOFLEX & 300 \\

GELAFIX & 100 \\

GELATIN & 100 \\

GLUE & 200  \\

KNETOSIL & 200 \\

LATEX & 100 \\

MOULDABLE-CLAY & 100 \\

MOULDABLE-GLUE & 900 \\

PAPER PRINTOUT & 1200 \\

PLAYDOH & 1700 \\

SILLY-PUTTY & 600 \\
\hline
\end{tabular}
\caption{Statisitcs of the COLFISPOOF dataset\cite{kolberg2023colfispoof}}
\label{COLFISPOOFDATASET}
\end{table}

\subsection{Metrics and Evaluation Protocol}
For testing the algorithm on both CLARKSON and COLFISPOOF database, three metrics has been defined and used, bonafide presentation classification error rate (BPCER) refers to the proportion of genuine presentations that are incorrectly classified as attack presentations, attack presentation classification error rate (APCER) refers to the proportion of spoofed image that are incorrectly classified as bonafide presentation, and average classification error rate (ACER) which refers to average of APCER and BPCER for the comparison. Moreover, we utilized receiver operating characteristic (ROC) curve see the performance of our model in classification between live and spoof data samples. As we discussed in previous sections, we implemented unsupervised training, exclusively using the CLARKSON live dataset to train the model. Subsequently, we evaluated the model's performance using both the Live dataset (CLARKSON) and the SPOOF dataset (CLARKSON, CLFISPOOF).
% Additionally, we employed k-fold cross-validation to assess the model's effectiveness on both live and spoof datasets. The primary objective of k-fold cross-validation is to evaluate the performance of the trained model on unseen live subjects. For training, we utilized two methods. 
Firstly, we split the live dataset into training and testing sets, ensuring that all the live subjects were included in both sets. Secondly, we implemented K-fold cross-validation to estimate the performance of our model on unseen live subjects~\cite{berrar2019cross}. In each fold, we allocated 80\% of the subjects for training and 20\% for testing. In the subsequent sections, we will present the results for each training method.

\subsection{Experimental Results}
\label{result}

\begin{figure}
    \centering
    \includegraphics[width=1\linewidth]{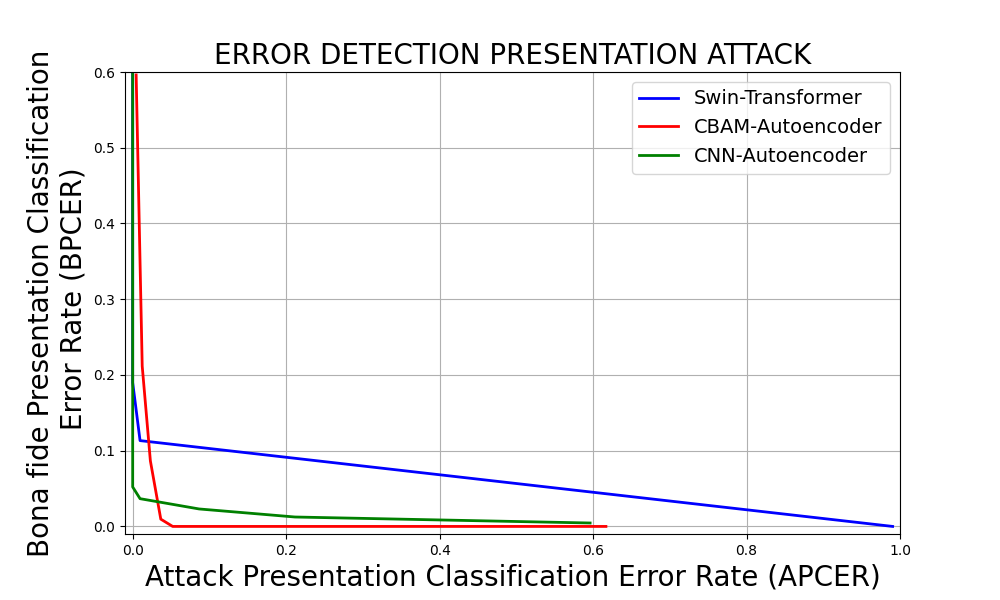}
    \caption{ROC curve of three different deep learning approach for PAD}
    \label{fig:Error}
\end{figure}

We have implemented three methods including, CAE, CAAE, and swin transformer for contactless fingerprint anti-spoofing detection purpose. CAE, contains 5 convolution layers in both encoder and decoder, where each convolution layer consist of rectified linear unit (ReLU) activation function followed by sigmoid function in the last layer of decoder. We set the kernel size to 4, with stride = 2. We set the live input image size to 256 $\times$ 256 with 3 channels (RGB), and reduce the dimension to latent representation 4 $\times$ 4 with 2048 channel. Finally, the decoder reconstruct the image with same size as input live image. To evaluate the model we use both live and spoof dataset (See Table.~\ref{tab:finalresult}, for the convolutional autoencoder results). CAE is a proper architecture to learn the representation of input live data, however, in order to improve the classification accuracy, we applied attention mechanism to the convolutional autoencoder. As we discussed in section~\ref{sec:Proposed}, we applied channel attention and spatial attention on convolutional autoencoder to capture the inter-dependencies between different channels of the live input feature maps and focusing on capturing the inter-dependencies between spatial locations within each channel of the feature maps and finally improve the model performance in classification. We incorporated the swin transformer into our classification approach to enhance its capabilities. Based on the data presented in Table ~\ref{tab:finalresult}, our CBAM-autoencoder model demonstrates superior performance across various types of spoof samples from CLARKSON, yielding an average APCER of 1.6\%. Notably, for COLFISPOOFED and IIITD samples, the APCER is 0\%. Additionally, the BPCER is measured at 0.96\%, contributing to an ACER value of 1.28\%, outperforming the two alternative approaches. Finally, in Figure.~\ref{fig:Error} we plot bonafide presentation classification error rate (BPCER) to attack presentation classification error rate (APCER) to shows the error rate of three implemented methods. According to Figure.~\ref{fig:Error}, the CBAM-autoencoder error rate reduced than the convolutional autoencoder and swin transformer.

\begin{table}[htb]
\centering
\begin{tabular}{ccc}
\hline
Fold\_number  & BPCER(\%) & TPR(\%)  \\
\hline
Fold\_1      & 3.7   & 96.28   \\
Fold\_2      & 1.9   & 98.56     \\
Fold\_3      & 1.3   & 98.68    \\
Fold\_4      & 4.1  & 95.87  \\
Fold\_5      & 2.03  & 97   \\
\hline
\end{tabular}
\caption{K-fold Cross Validation on Live CLARKSON dataset }
\label{tab:kfold}
\end{table}

\begin{table*}[htbp]
\centering
\resizebox{\textwidth}{!}{%
\begin{tabular}{|llccccccccccccccccccc|}
\hline
\multicolumn{21}{|c|}{ClARKSON} \\ \hline
\multicolumn{2}{|l|}{\multirow{2}{*}{Method}} & \multicolumn{1}{l|}{\multirow{2}{*}{DL}} & \multicolumn{2}{c|}{\begin{tabular}[c]{@{}c@{}}TRAINING\\ SAMPLES\end{tabular}} & \multicolumn{2}{c|}{\begin{tabular}[c]{@{}c@{}}TESTING\\ SAMPLES\end{tabular}} & \multicolumn{12}{c|}{APCER\%} & \multicolumn{1}{c|}{\multirow{2}{*}{\begin{tabular}[c]{@{}c@{}}BPCER\\ \%\end{tabular}}} & \multirow{2}{*}{\begin{tabular}[c]{@{}c@{}}HTER\\ \%\end{tabular}} \\ \cline{4-19}
\multicolumn{2}{|l|}{} & \multicolumn{1}{l|}{} & \multicolumn{1}{l|}{LIVE} & \multicolumn{1}{l|}{SPOOF} & \multicolumn{1}{l|}{LIVE} & \multicolumn{1}{l|}{SPOOF} & \multicolumn{3}{c|}{ECO} & \multicolumn{3}{c|}{PH} & \multicolumn{3}{c|}{PL} & \multicolumn{3}{c|}{WO} & \multicolumn{1}{c|}{} &  \\ \hline
\multicolumn{2}{|l|}{DENSENET-121~\cite{purnapatra2023presentation}} & \multicolumn{1}{c|}{S} & \multicolumn{2}{c|}{55,624} & \multicolumn{2}{c|}{5,590} & \multicolumn{3}{c|}{0} & \multicolumn{3}{c|}{88.03} & \multicolumn{3}{c|}{0.14} & \multicolumn{3}{c|}{0} & \multicolumn{1}{c|}{0.18} & NA \\ 
\multicolumn{2}{|l|}{DENSENET-121(KERAS)~\cite{purnapatra2023presentation}} & \multicolumn{1}{c|}{S} & \multicolumn{2}{c|}{55,624} & \multicolumn{2}{c|}{5,590} & \multicolumn{3}{c|}{0} & \multicolumn{3}{c|}{79.01} & \multicolumn{3}{c|}{1.55} & \multicolumn{3}{c|}{0.94} & \multicolumn{1}{c|}{3.64} & NA \\ 
\multicolumn{2}{|l|}{NASNETMOBILE~\cite{purnapatra2023presentation}} & \multicolumn{1}{c|}{S} & \multicolumn{2}{c|}{55,624} & \multicolumn{2}{c|}{5,590} & \multicolumn{3}{c|}{0} & \multicolumn{3}{c|}{82.15} & \multicolumn{3}{c|}{0.71} & \multicolumn{3}{c|}{5.96} & \multicolumn{1}{c|}{9.04} & NA \\ 
\multicolumn{2}{|l|}{DENSENET-121(GRAYSCALE)~\cite{purnapatra2023presentation}} & \multicolumn{1}{c|}{S} & \multicolumn{2}{c|}{55,624} & \multicolumn{2}{c|}{5,590} & \multicolumn{3}{c|}{0.16} & \multicolumn{3}{c|}{98.9} & \multicolumn{3}{c|}{1.98} & \multicolumn{3}{c|}{11} & \multicolumn{1}{c|}{0.18} & NA \\ \hline
\multicolumn{1}{|l|}{\multirow{3}{*}{CASE-1}} & \multicolumn{1}{l|}{RESNET50~\cite{li2023deep}} & \multicolumn{1}{c|}{S} & \multicolumn{1}{c|}{NA} & \multicolumn{1}{c|}{2099} & \multicolumn{1}{c|}{NA} & \multicolumn{1}{c|}{1248} & \multicolumn{3}{c|}{13.32} & \multicolumn{3}{c|}{-} & \multicolumn{3}{c|}{-} & \multicolumn{3}{c|}{-} & \multicolumn{1}{c|}{3.33} & 6.9 \\ 
\multicolumn{1}{|l|}{} & \multicolumn{1}{l|}{DENSNET201~\cite{li2023deep}} & \multicolumn{1}{c|}{S} & \multicolumn{1}{c|}{NA} & \multicolumn{1}{c|}{2099} & \multicolumn{1}{c|}{NA} & \multicolumn{1}{c|}{1248} & \multicolumn{3}{c|}{15.33} & \multicolumn{3}{c|}{--} & \multicolumn{3}{c|}{-} & \multicolumn{3}{c|}{-} & \multicolumn{1}{c|}{3.33} & 7.61 \\
\multicolumn{1}{|l|}{} & \multicolumn{1}{l|}{EFFICIENTNET-B0~\cite{li2023deep}} & \multicolumn{1}{c|}{S} & \multicolumn{1}{c|}{NA} & \multicolumn{1}{c|}{2099} & \multicolumn{1}{c|}{NA} & \multicolumn{1}{c|}{1248} & \multicolumn{3}{c|}{16.72} & \multicolumn{3}{c|}{-} & \multicolumn{3}{c|}{-} & \multicolumn{3}{c|}{-} & \multicolumn{1}{c|}{3.33} & 8.07 \\ \hline
\multicolumn{1}{|l|}{\multirow{3}{*}{CASE-2}} & \multicolumn{1}{l|}{RESNET50~\cite{li2023deep}} & \multicolumn{1}{c|}{S} & \multicolumn{1}{c|}{NA} & \multicolumn{1}{c|}{3143} & \multicolumn{1}{c|}{NA} & \multicolumn{1}{c|}{1104} & \multicolumn{3}{c|}{-} & \multicolumn{3}{c|}{45.33} & \multicolumn{3}{c|}{-} & \multicolumn{3}{c|}{-} & \multicolumn{1}{c|}{3.33} & 17.61 \\ 
\multicolumn{1}{|l|}{} & \multicolumn{1}{l|}{DENSNET201~\cite{li2023deep}} & \multicolumn{1}{c|}{S} & \multicolumn{1}{c|}{NA} & \multicolumn{1}{c|}{3143} & \multicolumn{1}{c|}{NA} & \multicolumn{1}{c|}{1104} & \multicolumn{3}{c|}{-} & \multicolumn{3}{c|}{94.72} & \multicolumn{3}{c|}{-} & \multicolumn{3}{c|}{-} & \multicolumn{1}{c|}{3.33} & 34.07 \\ 
\multicolumn{1}{|l|}{} & \multicolumn{1}{l|}{EFFICIENTNET-B0~\cite{li2023deep}} & \multicolumn{1}{c|}{S} & \multicolumn{1}{c|}{NA} & \multicolumn{1}{c|}{3143} & \multicolumn{1}{c|}{NA} & \multicolumn{1}{c|}{1104} & \multicolumn{3}{c|}{-} & \multicolumn{3}{c|}{100} & \multicolumn{3}{c|}{-} & \multicolumn{3}{c|}{-} & \multicolumn{1}{c|}{3.33} & \multicolumn{1}{l|}{38.89} \\ \hline
\multicolumn{1}{|l|}{\multirow{3}{*}{CASE-3}} & \multicolumn{1}{l|}{RESNET50~\cite{li2023deep}} & \multicolumn{1}{c|}{S} & \multicolumn{1}{c|}{NA} & \multicolumn{1}{c|}{2624} & \multicolumn{1}{c|}{NA} & \multicolumn{1}{c|}{1623} & \multicolumn{3}{c|}{-} & \multicolumn{3}{c|}{-} & \multicolumn{3}{c|}{17.34} & \multicolumn{3}{c|}{-} & \multicolumn{1}{c|}{3.33} & 8.28 \\  
\multicolumn{1}{|l|}{} & \multicolumn{1}{l|}{DENSNET201~\cite{li2023deep}} & \multicolumn{1}{c|}{S} & \multicolumn{1}{c|}{NA} & \multicolumn{1}{c|}{2624} & \multicolumn{1}{c|}{NA} & \multicolumn{1}{c|}{1623} & \multicolumn{3}{c|}{-} & \multicolumn{3}{c|}{-} & \multicolumn{3}{c|}{100} & \multicolumn{3}{c|}{-} & \multicolumn{1}{c|}{3.33} & 50 \\ 
\multicolumn{1}{|l|}{} & \multicolumn{1}{l|}{EFFICIENTNET-B0~\cite{li2023deep}} & \multicolumn{1}{c|}{S} & \multicolumn{1}{c|}{NA} & \multicolumn{1}{c|}{2624} & \multicolumn{1}{c|}{NA} & \multicolumn{1}{c|}{1623} & \multicolumn{3}{c|}{-} & \multicolumn{3}{c|}{-} & \multicolumn{3}{c|}{100} & \multicolumn{3}{c|}{-} & \multicolumn{1}{c|}{3.33} & 38.89 \\ \hline
\multicolumn{1}{|l|}{\multirow{3}{*}{CASE-4}} & \multicolumn{1}{l|}{RESNET50~\cite{li2023deep}} & \multicolumn{1}{c|}{S} & \multicolumn{1}{c|}{NA} & \multicolumn{1}{c|}{3975} & \multicolumn{1}{c|}{NA} & \multicolumn{1}{c|}{272} & \multicolumn{3}{c|}{-} & \multicolumn{3}{c|}{-} & \multicolumn{3}{c|}{-} & \multicolumn{3}{c|}{0.39} & \multicolumn{1}{c|}{3.33} & 2.63 \\ 
\multicolumn{1}{|l|}{} & \multicolumn{1}{l|}{DENSNET201~\cite{li2023deep}} & \multicolumn{1}{c|}{S} & \multicolumn{1}{c|}{NA} & \multicolumn{1}{c|}{2975} & \multicolumn{1}{c|}{NA} & \multicolumn{1}{c|}{272} & \multicolumn{3}{c|}{-} & \multicolumn{3}{c|}{-} & \multicolumn{3}{c|}{-} & \multicolumn{3}{c|}{5.97} & \multicolumn{1}{c|}{3.33} & 4.49 \\ 
\multicolumn{1}{|l|}{} & \multicolumn{1}{l|}{EFFICIENTNET-B0~\cite{li2023deep}} & \multicolumn{1}{c|}{S} & \multicolumn{1}{c|}{NA} & \multicolumn{1}{c|}{2975} & \multicolumn{1}{c|}{NA} & \multicolumn{1}{c|}{272} & \multicolumn{3}{c|}{-} & \multicolumn{3}{c|}{-} & \multicolumn{3}{c|}{-} & \multicolumn{3}{c|}{1.74} & \multicolumn{1}{c|}{3.33} & 3.08 \\ \hline
\multicolumn{2}{|l|}{\textbf{AUTOENCODER+CBAM (OURS)}} & \multicolumn{1}{c|}{\textbf{U}} & \multicolumn{1}{c|}{\textbf{4208}} & \multicolumn{1}{c|}{\textbf{0}} & \multicolumn{1}{c|}{\textbf{1678}} & \multicolumn{1}{c|}{\textbf{4324}} & \multicolumn{3}{c|}{\textbf{0}} & \multicolumn{3}{c|}{\textbf{29.35}} & \multicolumn{3}{c|}{\textbf{0}} & \multicolumn{3}{c|}{\textbf{0}} & \multicolumn{1}{c|}{\textbf{0.96}} & \textbf{3.43} \\ 
\multicolumn{2}{|l|}{AUTOENCODER(OURS)} & \multicolumn{1}{c|}{U} & \multicolumn{1}{c|}{4208} & \multicolumn{1}{c|}{0} & \multicolumn{1}{c|}{1678} & \multicolumn{1}{c|}{4324} & \multicolumn{3}{c|}{0} & \multicolumn{3}{c|}{67.3} & \multicolumn{3}{c|}{0} & \multicolumn{3}{c|}{0.73} & \multicolumn{1}{c|}{1.92} & 4.79 \\ 
\multicolumn{2}{|l|}{SWIN TRANSFORMER(OURS)} & \multicolumn{1}{c|}{U} & \multicolumn{1}{c|}{4208} & \multicolumn{1}{c|}{0} & \multicolumn{1}{c|}{1678} & \multicolumn{1}{c|}{4324} & \multicolumn{3}{c|}{37.1} & \multicolumn{3}{c|}{73.28} & \multicolumn{3}{c|}{0} & \multicolumn{3}{c|}{1.11} & \multicolumn{1}{c|}{0.96} & 5.9 \\ \hline
\multicolumn{20}{|c|}{COLFISPOOOF} &  \\ \hline
\multicolumn{2}{|l|}{\multirow{2}{*}{METHOD}} & \multicolumn{1}{l|}{\multirow{2}{*}{DL}} & \multicolumn{2}{c|}{\begin{tabular}[c]{@{}c@{}}TRAINING\\ SAMPLES\end{tabular}} & \multicolumn{2}{c|}{\begin{tabular}[c]{@{}c@{}}TESTING\\ SAMPLES\end{tabular}} & \multicolumn{12}{c|}{APCER\%} & \multicolumn{1}{c|}{\multirow{2}{*}{\begin{tabular}[c]{@{}c@{}}BPCER\\ \%\end{tabular}}} & \multirow{2}{*}{\begin{tabular}[c]{@{}c@{}}HTER\\ \%\end{tabular}} \\ \cline{4-19}
\multicolumn{2}{|l|}{} & \multicolumn{1}{l|}{} & \multicolumn{1}{l|}{LIVE} & \multicolumn{1}{l|}{SPOOF} & \multicolumn{1}{l|}{LIVE} & \multicolumn{1}{l|}{SPOOF} & \multicolumn{1}{l|}{ECO} & \multicolumn{1}{l|}{GEL1} & \multicolumn{1}{l|}{GEL2} & \multicolumn{1}{l|}{GLUE1} & \multicolumn{1}{l|}{KNE} & \multicolumn{1}{l|}{LA} & \multicolumn{1}{l|}{CLAY} & \multicolumn{1}{c|}{GLUE2} & \multicolumn{1}{l|}{PA} & \multicolumn{1}{l|}{PL} & \multicolumn{1}{l|}{SL} & \multicolumn{1}{l|}{DR} & \multicolumn{1}{c|}{} &  \\ \hline
\multicolumn{2}{|l|}{AUTOENCODER(OURS)} & \multicolumn{1}{c|}{U} & \multicolumn{1}{c|}{4208} & \multicolumn{1}{c|}{0} & \multicolumn{1}{c|}{1678} & \multicolumn{1}{c|}{7200} & \multicolumn{1}{c|}{0} & \multicolumn{1}{c|}{0} & \multicolumn{1}{c|}{0} & \multicolumn{1}{c|}{0} & \multicolumn{1}{c|}{0} & \multicolumn{1}{c|}{0} & \multicolumn{1}{c|}{0} & \multicolumn{1}{c|}{0} & \multicolumn{1}{c|}{0} & \multicolumn{1}{c|}{0} & \multicolumn{1}{c|}{0} & \multicolumn{1}{c|}{0} & \multicolumn{1}{c|}{1.92} &  0.96\\ 
\multicolumn{2}{|l|}{\textbf{AUTOENCODER+CBAM (OURS)}} & \multicolumn{1}{c|}{\textbf{U}} & \multicolumn{1}{c|}{\textbf{4208}} & \multicolumn{1}{c|}{\textbf{0}} & \multicolumn{1}{c|}{\textbf{1678}} & \multicolumn{1}{c|}{\textbf{7200}} & \multicolumn{1}{c|}{\textbf{0}} & \multicolumn{1}{c|}{\textbf{0}} & \multicolumn{1}{c|}{\textbf{0}} & \multicolumn{1}{c|}{\textbf{0}} & \multicolumn{1}{c|}{\textbf{0}} & \multicolumn{1}{c|}{\textbf{0}} & \multicolumn{1}{c|}{\textbf{0}} & \multicolumn{1}{c|}{\textbf{0}} & \multicolumn{1}{c|}{\textbf{0}} & \multicolumn{1}{c|}{\textbf{0}} & \multicolumn{1}{c|}{\textbf{0}} & \multicolumn{1}{c|}{\textbf{0}} & \multicolumn{1}{c|}{\textbf{0.96}} & \textbf{0.48} \\ 
\multicolumn{2}{|l|}{SWIN TRANSFORMER(OURS)} & \multicolumn{1}{c|}{U} & \multicolumn{1}{c|}{4208} & \multicolumn{1}{c|}{0} & \multicolumn{1}{c|}{1678} & \multicolumn{1}{c|}{7200} & \multicolumn{1}{c|}{0} & \multicolumn{1}{c|}{0} & \multicolumn{1}{c|}{0} & \multicolumn{1}{c|}{0} & \multicolumn{1}{c|}{0} & \multicolumn{1}{c|}{0} & \multicolumn{1}{c|}{0} & \multicolumn{1}{c|}{0} & \multicolumn{1}{c|}{0} & \multicolumn{1}{c|}{0} & \multicolumn{1}{c|}{0} & \multicolumn{1}{c|}{0} & \multicolumn{1}{c|}{0.96} &0.48  \\ \hline
\multicolumn{21}{|c|}{IIITD DATASET} \\ \hline
\multicolumn{2}{|c|}{\multirow{2}{*}{METHOD}} & \multicolumn{1}{c|}{\multirow{2}{*}{DL}} & \multicolumn{2}{c|}{\begin{tabular}[c]{@{}c@{}}TRAINING\\ SAMPLES\end{tabular}} & \multicolumn{2}{c|}{\begin{tabular}[c]{@{}c@{}}TESTING\\ SAMPLES\end{tabular}} & \multicolumn{12}{c|}{\begin{tabular}[c]{@{}c@{}}APCER\\ \%\end{tabular}} & \multicolumn{1}{c|}{\multirow{2}{*}{\begin{tabular}[c]{@{}c@{}}BPCER\\ \%\end{tabular}}} & \multirow{2}{*}{\begin{tabular}[c]{@{}c@{}}HTER\\ \%\end{tabular}} \\ \cline{4-19}
\multicolumn{2}{|c|}{} & \multicolumn{1}{c|}{} & \multicolumn{1}{c|}{LIVE} & \multicolumn{1}{c|}{SPOOF} & \multicolumn{1}{c|}{LIVE} & \multicolumn{1}{c|}{SPOOF} & \multicolumn{6}{c|}{PA} & \multicolumn{6}{c|}{PH} & \multicolumn{1}{c|}{} &  \\ \hline
\multicolumn{2}{|l|}{ALEXNET~\cite{fujio2018face}} & \multicolumn{1}{c|}{S} & \multicolumn{1}{c|}{NA} & \multicolumn{1}{c|}{NA} & \multicolumn{1}{c|}{NA} & \multicolumn{1}{c|}{NA} & \multicolumn{6}{c|}{NA} & \multicolumn{6}{c|}{NA} & \multicolumn{1}{c|}{NA} & 0.04 \\ 
\multicolumn{2}{|l|}{RESNET18~\cite{marasco2021fingerphoto}} & \multicolumn{1}{c|}{S} & \multicolumn{1}{c|}{NA} & \multicolumn{1}{c|}{NA} & \multicolumn{1}{c|}{NA} & \multicolumn{1}{c|}{NA} & \multicolumn{6}{c|}{NA} & \multicolumn{6}{c|}{NA} & \multicolumn{1}{c|}{NA} & NA \\ 
\multicolumn{2}{|l|}{DENSENET121~\cite{marasco2021deep}} & \multicolumn{1}{c|}{S} & \multicolumn{1}{c|}{NA} & \multicolumn{1}{c|}{NA} & \multicolumn{1}{c|}{NA} & \multicolumn{1}{c|}{NA} & \multicolumn{6}{c|}{NA} & \multicolumn{6}{c|}{NA} & \multicolumn{1}{c|}{NA} & 1.274 \\ 
\multicolumn{2}{|l|}{\textbf{AUTOENCODER+CBAM(OURS)}} & \multicolumn{1}{c|}{\textbf{U}} & \multicolumn{1}{c|}{\textbf{4208}} & \multicolumn{1}{c|}{\textbf{0}} & \multicolumn{1}{c|}{\textbf{1678}} & \multicolumn{1}{c|}{\textbf{8192}} & \multicolumn{6}{c|}{\textbf{0}} & \multicolumn{6}{c|}{\textbf{0}} & \multicolumn{1}{c|}{\textbf{0.96}} & \textbf{0.48} \\ 
\multicolumn{2}{|l|}{AUTOENCODER(OURS)} & \multicolumn{1}{c|}{U} & \multicolumn{1}{c|}{4208} & \multicolumn{1}{c|}{0} & \multicolumn{1}{c|}{1678} & \multicolumn{1}{c|}{8192} & \multicolumn{6}{c|}{0} & \multicolumn{6}{c|}{0.91} & \multicolumn{1}{c|}{1.92} & 1.13 \\ 
\multicolumn{2}{|l|}{SWIN TRANSFORMER(OURS)} & \multicolumn{1}{c|}{U} & \multicolumn{1}{c|}{4208} & \multicolumn{1}{c|}{0} & \multicolumn{1}{c|}{1678} & \multicolumn{1}{c|}{8192} & \multicolumn{6}{c|}{0} & \multicolumn{6}{c|}{43.11} & \multicolumn{1}{c|}{0.96} & 16.48 \\ \hline
\end{tabular}%
}
\caption{Comparison of proposed method with previous works in terms of APCER and BPCER. DL--deep learning techniques which S is supervised and U is unsupervised, NA--- not reported. ECO--ecoflex, PH--photo paper, PA-- printed attack, PL--playdoh, WO--woodglue, DR--dragonskin, GEL1--gelatin, GEL2--gelafix, GLUE1--GLUE, KNE--knetosil, LA--latex, CLAY--mouldableclay, GLUE2--modulableglue, SL--silly putty.}
\label{tab:finalresult}
\end{table*}
We initially anticipated the swin transformer will improve results with this approach. However, as indicated in Table.~\ref{tab:finalresult}, the performance of the swin transformer in detecting spoof datasets (photopaper and woodglue) in the CLARKSON dataset does not surpass that of the CBAM-autoencoder. This outcome may be attributed to the limited quality and quantity of the training dataset. However, the bonafide classification error rate of the swin transformer improved by 1\% compared to CBAM-Autoencoder and 50\% compared to convolutional autoencoder. Based on our finding, the swin transformer successfully detect all the different types of spoofs on the CLFISPOOF dataset as ``spoof," achieving an APCER of 0\%.

\subsection{Discussion and Comparison}
Based on the results demonstrated in Table~\ref{tab:finalresult}, our proposed methods Auto-CBAM achieved the best performance compared to state of art techniques~\cite{purnapatra2023presentation,li2023deep}. In order to contrast our approach with previous research, various metrics are considered, including supervised versus unsupervised methodologies, differences in training and testing sample sizes, and variations in the types of spoofed samples examined during testing. To begin with, both the studies by Purnapatra et al.~\cite{purnapatra2023presentation} (2023) and Li et al.~\cite{li2023deep} (2023) are utilized supervised techniques, while our approach adopts an unsupervised framework. This distinction signifies that their models were trained using both spoofed and live samples, whereas in our methodology, we exclusively employ live samples for training purposes. Furthermore, our evaluation diverges from conventional approaches by assessing the performance on previously unseen authentic subjects when calculating the BPCER. In contrast, the previous works utilize unseen samples from trained subjects for their assessment. sing two distinct databases, marking the first instance of such an assessment. In contrast, existing work are only evaluated based on the Clarkson database. As depicted in Table~\ref{tab:finalresult}, our model's evaluation includes 11,524 spoofed samples. This number significantly exceeds the quantity of spoofed samples employed in the study by Li et al. (2023) by approximately 10,000 and surpasses the count in the research conducted by Purnapatra et al. (2023) by an additional 7,500 spoofed samples. Based on our analysis, we found photo paper is most challenging spoofed samples compared to other types of spoofed samples. Having said that, our model successfully shows a degradation of the error rate of the APCER by 49.66\% when compared to the performance of DenseNet-121 (Keras). Furthermore, it demonstrates a 58.68\% reduction in error rate in comparison to standard DenseNet-121, a 69.55\% decrease when contrasted with DenseNet-121 trained on grayscale data, and a 52.8\% improvement over NasNetMobile, as cited in Purnapatra et al.'s work~\cite{purnapatra2023presentation}. Also, our proposed method performance in detecting photo paper as spoof improved by 15.98\% compared with ResNet50, 65.37\% compared with DenseNet201, 70.65\% compared with EfficientNet-BO in~\cite{li2023deep}. Note that BPCER and APCER from Li et al.~\cite{li2023deep} were adjusted for better comparison. We also compared our work with \cite{fujio2018face,marasco2021fingerphoto,marasco2021deep} using IIITD spoofed. Similar to other comparison, we only trained our model using live samples from CLARKSON database and evaluated under IIITD spoofed samples. As we described in datasets, the IIITD dataset comprises two spoof samples including 2,048 images from print attacks, and 6,144 images from photo attacks. As can be seen in Table~\ref{tab:finalresult}, our model achieved ``zero'' of APCER on both printed and photo attacks while existing work did not reported APCER. It is worth noting that, unlike prior studies that omitted reporting the APCER, we have explicitly included APCER reporting in table~\ref{tab:finalresult}. We also noticed that, while our proposed Auto-CBAM and autoencoder architecture work very well on photo paper, swing transformer performance exhibits degradation. one reason is that, by incorporating both channel-wise and spatial-wise attention mechanisms in CBAM, where both global contextual information (channel-wise) and fine-grained local details (spatial-wise) is captured will help to outperformed compared to other model. 

We also illustrated that swin transformer resulted 73.63\% improvement on the BPCER compared to DenseNet-121 (keras)~\cite{purnapatra2023presentation}. While the Swin Transformer did not outperform CBAM-Autoencoder, it still exhibited the APCER improved by 16.7\% compared to DenseNet-121, 7.25\% compared to DenseNet-121 (keras), 10.79\% compared to NasNetMobile, and 25.9\% compared to DenseNet-121 (grayscale) for photo paper attacks. We also implement K-fold cross-validation to evaluate the performance of CBAM-Autoencoder model on unseen live subject data. Table.~\ref{tab:kfold} demonstrates the performance of CBAM-Autoencoder with K-fold cross validation training. As we mentioned, the CLARKSON dataset, contains 26 subjects for live dataset. For k-fold cross validation implementation, we allocated 20\% of the unseen subjects (4 Subjects) for testing and the remaining used for training to evaluate the performance of our model on unseen live subjects. According to the Table.~\ref{tab:finalresult}, our model achieved performed well in each fold, indicating that our model is capable of classifying unseen live subjects as live.

\section{Conclusion}
%\vspace{-8pt}
\label{sec:conclude}
%Contactless fingerprint biometric system is rapidly gaining popularity and holds the potential to be replaced by conventional touch-based fingerprint biometric recognition systems. However, there are several drawbacks such as presentation attack such as photo paper and paper printout attack.
Current research on presentation attack detection (PAD) primarily relies on supervised learning techniques, where both bonafide samples and spoofed samples are utilized during training which are not scalable due to poor performance against unseen attack. In this paper, we introduced PAD using unsupervised approach that combines an unsupervised autoencoder with a convolutional block attention. Our proposed deep learning approach is only trained on bonafide images without exposure to any spoofed samples. It is then evaluated against unseen spoofed samples in the testing phase. The scheme we proposed has achieved an average BPCER of 0.96\% with an APCER of 1.6\% for presentation attacks with various types of spoofed samples.

\section{Acknowledgments}
This project was supported in part by the National Science Foundation under Grants No. 2104520.

% 

% {\small
% \bibliographystyle{ieeetr}
% \bibliography{egbib}
% }

\end{document}